\documentclass{article}
\usepackage{graphicx} 
\usepackage{subcaption}
\usepackage{booktabs}
\usepackage{caption}
\usepackage[square,numbers]{natbib}
\usepackage{ragged2e}

\usepackage[final]{neurips_2023}

\usepackage[utf8]{inputenc} % allow utf-8 input
\usepackage[T1]{fontenc}    % use 8-bit T1 fonts
\usepackage{hyperref}       % hyperlinks
\usepackage{url}            % simple URL typesetting
\usepackage{booktabs}       % professional-quality tables
\usepackage{amsfonts}       % blackboard math symbols
\usepackage{nicefrac}       % compact symbols for 1/2, etc.
\usepackage{microtype}      % microtypography
\usepackage{xcolor}         % colors

\title{Large Language Models Perform on Par with Experts Identifying Mental Health Factors in Adolescent Online Forums}

\author{%
  Isabelle Lorge
    \\
  Department of Psychiatry, University of Oxford, OX3 7JX\\
  \And
  Dan W. Joyce \\
  Department of Primary Care and Mental Health, University of Liverpool, L69 3GF \\
  \And
  Andrey Kormilitzin \\
  Department of Psychiatry, University of Oxford, OX3 7JX\\
}

\begin{document}

\maketitle

\begin{abstract}
Mental health in children and adolescents has been steadily deteriorating over the past few years \cite{o2020social}. The recent advent of Large Language Models (LLMs) offers much hope for cost and time efficient scaling of monitoring and intervention, yet despite specifically prevalent issues such as school bullying and eating disorders, previous studies on have not investigated performance in this domain or for open information extraction where the set of answers is not predetermined. We create a new dataset of Reddit posts from adolescents aged 12-19 annotated by expert psychiatrists for the following categories: TRAUMA, PRECARITY, CONDITION, SYMPTOMS, SUICIDALITY and TREATMENT and compare expert labels to annotations from two top performing LLMs (GPT3.5 and GPT4). In addition, we create two synthetic datasets to assess whether LLMs perform better when annotating data as they generate it. We find GPT4 to be on par with human inter-annotator agreement and performance on synthetic data to be substantially higher, however we find the model still occasionally errs on issues of negation and factuality and higher performance on synthetic data is driven by greater complexity of real data rather than inherent advantage.
\end{abstract}

\section{Introduction}

The recent development of powerful Large Language Models such as GPT3.5 \cite{brown2020language} and GPT4 \cite{achiam2023gpt}
able to perform tasks in a zero-shot manner (i.e., without having been specifically trained or fine-tuned to do so) by being simply prompted with natural language instructions shows much promise for healthcare applications and the domain of mental health. Indeed, these models display more impressive general natural language processing abilities than their predecessors  and excel at tasks such as Question Answering and Named Entity Recognition \cite{agrawal2022large, hu2023zero, liu2023deid, tang2023evaluating}. Models with the ability to process social media content for indicators of mental health issues have the potential to become invaluable cost-effective tools for applications such as public health monitoring \cite{graham2019artificial} and online moderation or intervention systems \cite{franco2023analyzing}. In addition, synthetic data produced by LLMs can be a cost effective and privacy-preserving tool for training task specific models \cite{lorge2024detecting}.

There have been several studies aimed at assessing the abilities of LLMs to perform a range of tasks related to mental health on datasets derived from social media. \citet{yang2023towards} conducted a comprehensive assessment of ChatGPT (gpt-3.5-turbo), InstructGPT3 and LlaMA7B and 13B \cite{touvron2023llama} on 11 different datasets and 5 tasks (mental health condition binary/multiclass detection, cause/factor detection, emotion detection and causal emotion entailment, i.e. determining the cause of a described emotion). They find that while the LLMs perform well (0.46-0.86 F1 depending on task), with ChatGPT substantially outperforming both LLaMA 7B and 13B, they still underperform smaller models specifically fine-tuned for each task (e.g., RoBERTa). \citet{xu2024mental} find similar results for Alpaca \cite{taori2023alpaca}, FLAN-T5 \cite{chung2022scaling} and LLaMA2 \cite{touvron2023llama2}, with only fine-tuned LLMs able to perform on par with smaller, task-specific models such as RoBERTa \cite{taylor2023clinical, taylor2024efficiency}. 

However, we find that previous studies suffer from the following shortcomings:
\begin{enumerate}
    \item They focus on adult mental health
    \item They focus on tasks with a closed (or finite) set of answers, where the model is asked to perform each task in turn
    \item They do not investigate how LLMs perform on synthetic data, i.e., text they are asked to simultaneously generate and label
\end{enumerate}

There is growing consensus that we are facing a child mental health crisis \cite{o2020social}. Before the COVID-19 pandemic there was already increasing incidence of mental health conditions in children and young people (CYP), such as depression, anxiety and eating disorders \cite{mojtabai2016national} as well as rising rates of self-harm and suicidal ideation \cite{marcheselli2018mental} and cyberbullying strongly linked to adverse mental health outcomes \cite{mishna2018social}. The advent of the pandemic accelerated this already precarious situation and created additional challenges \cite{orben2020effects, twenge2020us} such as discontinuity of healthcare service provision in addition to interruption to young people's usual engagement in education and their social lives. 

This age range is particularly vulnerable to onset of mental health issues, with half of conditions appearing by early adolescence and 10-20\% of children and young people experiencing at least one mental health condition \cite{kessler2007lifetime}. Females, those with low socioeconomic backgrounds, trauma, abuse or having witnessed violence \cite{rc2005lifetime} are at heightened risk.

On the other hand, social media now forms an important part of children and adolescents' daily lives, whose impact on mental health is debated, with potential benefits (stress reduction and support networks \cite{allen2014social}) as well as potential risks (sleep disturbance, self esteem issues and cyberbullying \cite{george2015seven}). 
Regardless of their detrimental or protective impact, social media may contribute valuable insights into CYP's mental health, with opportunities for monitoring and intervention, for example identifying those at risk of depression and mood disorders \cite{michikyan2020depression}. Given the mental health of CYP is a particularly pressing public health concern, we wished to investigate how LLMs perform on extracting mental health factors when faced with social media content generated by young people aged 12-19. Indeed, several issues related to mental health either exclusively apply to children and adolescents (such as school bullying and ongoing family abuse) or are particularly prevalent in this age range (such as eating disorders \cite{volpe2016eating} and self-harm \cite{griffin2018increasing}), making both content type and factors of interest distinct from those found in adult social media posts.

In addition, previous studies focused on tasks which had either a binary or closed sets of answers (e.g., choosing between several given conditions or between several given causal factors). In contrast, we wish to examine how LLMs perform on a task of open information extraction, where they are given categories of information and asked to extract any which are found in the text (e.g., asked to detect whether there is any mental health condition indicated in the text). Furthermore, in previous studies the models were tested with each task in turn (e.g., asked to detect depression in one dataset, then detect suicidality in another dataset), whereas we gather and annotate our own dataset in order to be able to ask the LLMs to extract all categories simultaneously (e.g, extract all conditions and symptoms in a given sentence). 

Finally, to our knowledge there has been no investigation on how LLM performance compares when asked to annotate text as they generate it, i.e., how their performance on synthetic data compares with their performance on real data. There is growing
interest in synthetic data for healthcare \cite{giuffre2023harnessing}. Given the potential for training
models and running simulations and digital twin experiments with the
benefit of reduced issues of data scarcity and privacy, we believe that our work will contribute to better understanding of limitations and benefits of using synthetic data for real-world tasks. 

\section{Aims}

In summary, we aim to:
\begin{enumerate}
    \item Generate and annotate with high-quality expert annotations a novel dataset of social media posts which allows extraction of a wide range of mental health factors simultaneously.
    \item Investigate performance of two top-performing LLMs (GPT3.5 and GPT4) on extracting mental health factors in adolescent social media posts to verify whether they can be on par with expert annotators.
    \item Investigate how these LLMs perform on synthetic data, i.e., when asked to annotate text as they generate it, with the aim of assessing utility of these data in training task specific models 
\end{enumerate}

\section{Method}
 
\subsection{Reddit dataset}
\begin{figure}[h!]
    \centering
    \includegraphics[width=\textwidth]{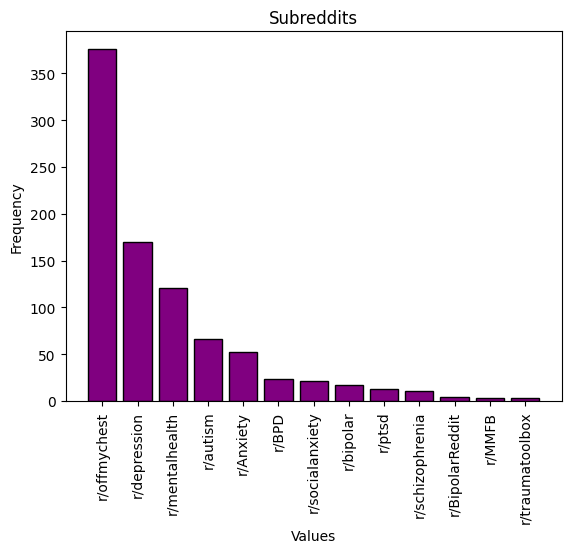}
    \caption{Distribution of subreddits}
    \label{fig:subreddits}
\end{figure}

\justifying{
We use Python's PRAW library to collect post from the Reddit website (www.reddit.com) over the last year, including posts from specific forum subthemes (`subreddits') dedicated to mental health topics: 
\textit{r/anxiety, r/depression, r/mentalhealth, r/bipolarreddit, r/bipolar, r/BPD, r/schizophrenia, r/PTSD, r/autism, r/trau-matoolbox, r/socialanxiety, r/dbtselfhelp, r/offmychest} and \textit{r/mmfb}. The distribution of subreddits in the dataset can be found in Figure \ref{fig:subreddits}.

As in previous works \cite{chew2021predicting}, we use heuristics to obtain posts from our target age range (e.g, posts containing expression such as \textit{I am 16/just turned 16/etc.}) We gather 1000 posts written by 950 unique users. To optimise the annotation process, we select the most relevant sentences to be annotated by embedding a set of mental health keywords with Python's \textit{sentence-transformers} library \cite{reimers-2019-sentence-bert} calculating the cosine similarity with post sentences, choosing a threshold of 0.2 cosine similarity after trial and error. We keep the post index for each sentence to provide context. The resulting dataset contains 6500 sentences.}

\subsection{Ethical considerations}
In conducting this research, we recognised the importance of respecting the autonomy and privacy of the Reddit users whose posts were included in our dataset. While Reddit data is publicly available and was obtained from open online forums, we acknowledge that users may not have anticipated their contributions being used for research purposes and will therefore make the data available only on demand. The verbatim example sentences given in later sections have been modified to prevent full-text searching strategies to infer the post author's immediate identity on reddit.

To protect the confidentiality of participants, we did not provide usernames or other identifying information to our annotators. Annotators were psychiatrists who were warned that the content of the posts was highly sensitive with potentially triggering topics such as self-harm and child abuse.

Reddit's data sharing and research policy allows academic researchers to access certain Reddit data for the purposes of research, subject to the platform's terms and conditions. They require researchers to obtain approval through their data access request process before using the API. The policy outlines requirements around protecting user privacy, obtaining consent, and properly attributing the data source in any published work. They reserve the right to deny data access requests or revoke access if the research is deemed to violate Reddit's policies. Researchers must also agree to Reddit's standard data use agreement when accessing the data.

Our research aims to contribute to the understanding of mental health discourse from adolescents on social media platforms. We believe the potential benefits of this work, in terms of insights that could improve mental health support and resources, outweigh the minimal risks to participants. However, we remain aware of the ethical complexities involved in using public social media data, and encourage further discussion and guidance in this emerging area of study.

\subsection{Synthetic dataset}
In addition to the real dataset, we generate two synthetic datasets of 500 sentences each by prompting GPT3.5 (\textit{gpt-3.5-turbo-0125}) and GPT4 (\textit{gpt-4-0125-preview}) to create and label Reddit-like posts of 5 sentences (temperature 0, all other parameters set to default). The instructions given were made as similar as possible to those given to annotators, and the model was expliclity told to only label factors which applied to the author of the post (e.g., not to label \textit{My friend has depression} with CONDITION). The prompt used can be found in Appendix \ref{appendixA}.

\subsection{Annotation schema}
\justifying{
Given our goal is to obtain a wide range of relevant annotations for each sentence in order to test the LLMs' ability to generalise and perform open information extraction, and the previously mentioned important factors related to trauma \cite{nelson2020adversity} and precarity \cite{fitzsimons2017poverty}, we create the following six categories in consultation with a clinical psychiatrist: 
\begin{itemize}
    \item \textbf{TRAUMA} (sexual abuse, physical abuse, emotional abuse, school bullying, death, accident, etc.)
    \item \textbf{PRECARITY} (socioeconomic, parental conflict, parental illness, etc.)
    \item \textbf{SYMPTOM} (self-harm, low self-esteem, anhedonia, panic attack, flashback, psychosis, insomnia, etc.)
    \item \textbf{CONDITION} (eating disorder, depression, bipolar, bpd, anxiety, ptsd, adhd, substance abuse/addiction, etc.)
    \item \textbf{SUICIDALITY} (no subcategories)
    \item \textbf{TREATMENT} (no subcategories)
\end{itemize}

Nineteen expert annotators  were contacted and asked to annotate 500 sentences each for a fixed compensation of £120 ($\approx \pounds60$/hour). These were UK-trained psychiatrists, all of whom had obtained Membership of the Royal College of Psychiatrists by post-graduate experience and formal examinations. Thirteen annotators annotated the Reddit dataset, two annotators annotated the synthetic datasets and four annotators re-annotated samples from the Reddit and synthetic datasets for inter-annotator agreement computation (100 sentences from each dataset, 1500 sentences in total). Annotators were given the above subcategory examples but allowed to use new subcategories when appropriate (no closed set of answers). They were given the post indices to provide context (i.e., so as to be aware which sentences belonged to the same post). They were asked to annotate only school bullying as bullying, and other instances (e.g., sibling harassment) as emotional abuse. Anxiety was to be annotated as a symptom rather than condition unless specifically described as a disorder.

Experts performed the annotation by filling in the relevant columns in an Excel sheet with each sentence as a row. Importantly, given the known limitations of language models with negation \cite{ettinger2020bert}, we wished to annotate both POSITIVE and NEGATIVE evidence in order to test LLMs' ability to handle both polarities (e.g., \textit{I am not feeling suicidal} as negative suicidality or \textit{We don't have any money issues} as negative socioeconomic precarity). For this purpose, annotators were asked to use the prefixes P and N (e.g., \textit{P(adhd)} in the CONDITION column or \textit{N(socioeconomic)} in the PRECARITY column).
}

\subsection{Data processing and dataset statistics}

In order to compare expert annotations with LLM annotations despite the wide variety of subcategories and terms used by annotators we create dictionaries mapping each term found in the dataset to a standard equivalent (e.g., \textit{p(emotional) to p(emotional abuse), p(physical violence) to p(physical abuse), p(gun violence) and p(school shooting) to p(violence), p(rape) to p(sexual abuse), p(financial burden) and p(poor) to p(socioeconomic precarity), p(divorce) to p(family conflict), p(self hatred) to p(low self esteem)}, etc.). Parental substance abuse is considered family illness and any underspecified subcategories are marked as `unspecified' (e.g., \textit{p(trauma unspecified)}). 

The distribution of subcategories for each category can be found in figures \ref{fig:trauma}, \ref{fig:precarity}, \ref{fig:condition} and \ref{fig:symptoms} in Appendix \ref{appendixB}. The most frequent subcategory in TRAUMA is emotional abuse, which occurs twice as often as physical abuse and death in the dataset. The most frequent form of PRECARITY is family conflict, then family illness (including parental substance abuse) and socioeconomic precarity. The most frequent CONDITIONS are depressive disorders, followed by substance abuse/addiction and ADHD. The most frequent SYMPTOMS are anxiety, low self-esteem, self-harm and low mood. 

Interestingly, the distribution of subcategories differs quite substantially in the synthetic datasets (distributions for the GPT3.5 and GPT4 generated datasets can be found in Appendix \ref{appendixB}). Overall, the number of subcategories is reduced, indicating less diversity (however, these are smaller datasets). The top trauma subcategories are sexual abuse for GPT3.5 and school bullying for GPT4, both of which were much less prevalent in real data. The second most prevalent condition for both GPT3.5 and GPT4 is eating disorders, whereas these ranked in 8th place in real data. Finally, unlike in real data, flashbacks and panic attacks are the 3d and 4th most frequent symptoms for both GPT3.5 and GPT4-generated data, whereas self-harm ranks much lower than in real data. Given many of these subcategories were given as examples in the annotator guidelines and LLM prompt, it is likely that the LLMs used them in a more homogenous manner for generation than the distribution which would be found in real data. However, the distribution is not entirely homogenous, which suggests the LLMs did leverage some of the biases learned from their training data. 

\section{Results}
Once both human and LLM annotations are standardised, we conduct analyses to assess performance. We provide precision, recall and F1 at the category level and accuracy at the subcategory level collapsed across subcategories (given their high number). We compute category performance in two ways: \textit{Positive or Negative}, where a point is awarded if the category contains an annotation in both human and LLM annotations, regardless of polarity (i.e., the annotator considered there was relevant information concerning the category TRAUMA) and \textit{Positive Only} metrics, where negative annotations are counted as no annotations. The difference between the two metrics can be seen clearly in Table \ref{gpt3_real} (GPT3.5 results), where precision increases but recall diminishes for \textit{Positive Only}. The increase in precision is due to the fact that GPT3.5 outputs a substantial number of negative annotations in cases where human annotators did not consider it relevant to mention the category. The reduction in recall, on the other hand, results from the fact that LLMs often confuse positive and negative annotations and will occasionally output a negative annotation for a positive one. 

For real data (Tables \ref{gpt3_real} and \ref{gpt4_real}), GPT3.5's performance at the category level is average, with better performance in the Positive Only metrics (0.57). GPT4 performs better, especially in Positive Only metrics (0.63) and subcategory accuracy (0.48 vs. 0.39). In general, recall is higher than precision, indicating LLMs may be overpredicting labels. 

The performance for synthetic data (Tables \ref{gpt3_synth} and \ref{gpt4_synth}) is substantially better, with no gap between the Positive or Negative and Positive Only metrics, suggesting less irrelevant negative annotations. Here again, GPT4 outperforms GPT3.5, both at the category level (0.75 vs 0.70 and 0.73 vs 0.68) and more particularly at the subcategory level, where GPT4 reaches an impressive accuracy of 0.72 (vs 0.42). The gap between recall and precision is reduced for GPT4, whereas GPT3.5 displays higher precision than recall here.

In order to assess the upper bound of human performance, we calculate inter-annotator agreement for both real and synthetic datasets using Cohen's Kappa. Values can be found in Table \ref{cohen}. Interestingly, while performance at the category level in real data is lower (GPT3.5) or similar (GPT4) compared to humans, GPT4 displays a substantially higher accuracy at the subcategory level (0.47 vs 0.35). For synthetic data, GPT3.5 still underperforms human agreement on all three metrics, while GPT4 is on par with humans for the Positive Only and subcategory metrics and only underperforms in the Positive and Negative metric.

\begin{table}[h!]
\centering
\footnotesize
\begin{tabular}{lrrrrrrrrr}
\toprule
Category & \multicolumn{3}{c}{Positive or Negative} & \multicolumn{4}{c}{Positive Only} & Subcategory \\
\cmidrule(l){2-9}
& Precision & Recall & F1-Score & & Precision & Recall & F1-Score & Accuracy\\
\midrule
TRAUMA & 0.38 & 0.78 & 0.51 & & 0.56 & 0.65 & 0.60 & 0.39\\
PRECARITY & 0.26 & 0.43 & 0.33 & & 0.45 & 0.31 & 0.37 & 0.22\\
CONDITION & 0.33 & 0.85 & 0.48 & & 0.54 & 0.72 & 0.62 & 0.55\\
SYMPTOMS & 0.39 & 0.62 & 0.48 & & 0.46 & 0.58 & 0.52 & 0.31\\
SUICIDALITY & 0.44 & 0.79 & 0.56 & & 0.80 & 0.68 & 0.73 & /\\
TREATMENT & 0.48 & 0.72 & 0.58 & & 0.72 & 0.58 & 0.64 & /\\
\midrule
ALL & 0.37 & 0.70 & 0.49 & & 0.55 & 0.60 & 0.57 & 0.39 \\
\bottomrule
\smallskip
\end{tabular}
\caption{GPT3.5 (real data). \textbf{Positive or Negative}: counting annotation in category regardless of polarity (category level); \textbf{Positive Only}: counting negative annotations as NaN (category level); \textbf{Subcategory}: accuracy at the subcategory level}
\label{gpt3_real}
\end{table}

\begin{table}[h!]
\centering
\footnotesize
\begin{tabular}{lrrrcrrrr}
\toprule
Category & \multicolumn{3}{c}{Positive or Negative} & \multicolumn{4}{c}{Positive Only} & Subcategory \\
\cmidrule(l){2-9}
& Precision & Recall & F1-Score & & Precision & Recall & F1-Score & Accuracy\\
\midrule
TRAUMA & 0.44 & 0.89 & 0.59 & & 0.57 & 0.84 & 0.68 & 0.57\\
PRECARITY & 0.31 & 0.52 & 0.39 & & 0.50 & 0.46 & 0.48 & 0.36 \\
CONDITION & 0.46 & 0.81 & 0.59 & & 0.61 & 0.77 & 0.68 & 0.57 \\
SYMPTOMS & 0.35 & 0.78 & 0.49 & & 0.45 & 0.73 & 0.56 & 0.41 \\
SUICIDALITY & 0.36 & 0.93 & 0.51 & & 0.70 & 0.87 & 0.77 & / \\
TREATMENT & 0.39 & 0.87 & 0.54 & & 0.64 & 0.81 & 0.71 & / \\
\midrule
ALL & 0.39 & 0.80 & 0.52 & & 0.55 & 0.75 & 0.63 & 0.48 \\
\bottomrule
\smallskip
\end{tabular}
\caption{GPT4 (real data). \textbf{Positive or Negative}: counting annotation in category regardless of polarity (category level); \textbf{Positive Only}: counting negative annotations as NaN (category level); \textbf{Subcategory}: accuracy at the subcategory level}
\label{gpt4_real}
\end{table}

\begin{table}[h!]
\centering
\footnotesize
\begin{tabular}{lrrrcrrrr}
\toprule
Category & \multicolumn{3}{c}{Positive or Negative} & \multicolumn{4}{c}{Positive Only} & Subcategory \\
\cmidrule(l){2-9}
& Precision & Recall & F1-Score & & Precision & Recall & F1-Score & Accuracy \\
\midrule
TRAUMA & 0.90 & 0.49 & 0.64 & & 0.90 & 0.49 & 0.64 & 0.38\\
PRECARITY & 0.84 & 0.69 & 0.76 & & 0.86 & 0.69 & 0.76 & 0.54 \\
CONDITION & 0.44 & 0.67 & 0.53 & & 0.47 & 0.67 & 0.55 & 0.59 \\
SYMPTOMS & 0.85 & 0.59 & 0.70 & & 0.84 & 0.59 & 0.69 & 0.36 \\
SUICIDALITY & 0.75 & 1.00 & 0.85 & & 0.77 & 0.90 & 0.83 & / \\
TREATMENT & 0.68 & 0.84 & 0.75 & & 0.76 & 0.57 & 0.65 & / \\
\midrule
ALL & 0.74 & 0.65 & 0.70 & & 0.77 & 0.61 & 0.68 & 0.42 \\
\bottomrule
\smallskip
\end{tabular}
\caption{GPT3.5 (synthetic data). \textbf{Positive or Negative}: counting annotation in category regardless of polarity (category level); \textbf{Positive Only}: counting negative annotations as NaN (category level); \textbf{Subcategory}: accuracy at the subcategory level}
\label{gpt3_synth}
\end{table}

\begin{table}[h!]
\centering
\footnotesize
\begin{tabular}{lrrrrrrrr}
\toprule
Category & \multicolumn{3}{c}{Positive or Negative} & \multicolumn{4}{c}{Positive Only} & Subcategory \\
\cmidrule(l){2-9}
& Precision & Recall & F1-Score & & Precision & Recall & F1-Score & Accuracy\\
\midrule
TRAUMA & 0.84 & 0.95 & 0.89 & & 0.86 & 0.92 & 0.89 & 0.82 \\
PRECARITY & 0.85 & 0.84 & 0.85 & & 0.91 & 0.82 & 0.86 & 0.80 \\
CONDITION & 0.61 & 0.67 & 0.64 & & 0.60 & 0.67 & 0.63 & 0.67 \\
SYMPTOMS & 0.49 & 0.78 & 0.60 & & 0.53 & 0.80 & 0.64 & 0.69 \\
SUICIDALITY & 0.81 & 0.94 & 0.87 & & 0.78 & 0.82 & 0.80 & / \\
TREATMENT & 0.85 & 0.89 & 0.87 & & 0.87 & 0.78 & 0.82 & / \\
\midrule
ALL & 0.69 & 0.83 & 0.75 & & 0.69 & 0.79 & 0.73 & 0.72 \\
\bottomrule
\smallskip
\end{tabular}
\caption{GPT4 (synthetic data). \textbf{Positive or Negative}: counting annotation in category regardless of polarity (category level); \textbf{Positive Only}: counting negative annotations as NaN (category level); \textbf{Subcategory}: accuracy at the subcategory level}
\label{gpt4_synth}
\end{table}

\begin{table}[h!]
    \centering
    \footnotesize
    \begin{tabular}{lrrr}
    \toprule
    & Positive and Negative & Positive Only & Subcategory \\
    \midrule
    Annotator vs. Annotator (real data) & 0.60 & 0.59 & 0.35 \\

    GPT3 vs. Annotator (real data) & 0.39 & 0.52 & 0.37 \\
    GPT4 vs. Annotator (real data) &  0.43 & 0.58 & 0.47 \\
    \midrule
    Annotator vs. Annotator (synthetic data) & 0.77 & 0.71 & 0.68 \\
    GPT3 vs. Annotator (synthetic data)  & 0.64 & 0.63 & 0.40 \\
    GPT4 vs. Annotator (synthetic data) &  0.70 & 0.69 & 0.71 \\
    \bottomrule
    \smallskip
    \end{tabular}
    \caption{Inter-annotator agreement (Cohen's Kappa)}
    \label{cohen}
\end{table}

\section{Error analysis}
We examine some of the sentences annotated by the LLMs in order to perform error analysis and extract the following findings (as mentioned previously some words have been paraphrased to preclude full-text search allowing user identification):
\begin{itemize}
    \item Both GPT3.5 and GPT4 produce infelicitous negations, i.e., negative annotations which would seem irrelevant to humans, e.g., (\textit{I have amazing people around me} =\textgreater  negative parental death or \textit{The internet is my one only coping mechanism} =\textgreater trauma unspecified)
    \item Despite being specifically prompted to only annotate factors related to the writer/speaker, LLMs (including GPT4) do not always comply, e.g., \textit{She comes from what is, honestly, a horrific family situation} =\textgreater emotional abuse)
    \item Even GPT4 makes errors regarding negation (e.g., \textit{I've read about people with autism getting temper tantrums/meltdowns, however, that has never really been a problem for me}=\textgreater negative autism or \textit{i had in my head that something inside was very wrong, but i never felt completely depressed all the time so i never took bipolar seriously} =\textgreater negative bipolar disorder)
    \item Despite being prompted to annotate suicidality in a separate category, LLMs often annotate it in the SYMPTOM rather than SUICIDALITY category
    \item GPT3.5 especially often outputs irrelevant/spurious/incorrect labels (e.g., `unemployed’ as condition, `ambition’ as symptom, labelling physical conditions instead of mental conditions only, etc.)
    \item Even GPT4 makes errors regarding factuality (e.g., \textit{It was around my second year in junior high school when my father tried to take his life} =\textgreater positive death)
\end{itemize}
However, in many cases the assessment is not entirely fair, as the LLMs (particularly GPT4) often catch annotations which human annotators missed, or the difference in subcategories is subjective and open to debate (e.g., school bullying vs emotional abuse, emotional abuse vs abuse unspecified, etc.). Thus it is possible that LLMs, or most likely GPT4, in fact outperformed experts on this task. 

\section{Discussion}
The results obtained from our comparison of LLM annotations with human annotations on both real and synthetic data allow us to make a few conclusions and recommendations. 

Overall, both LLMs perform well. Inter-annotator agreement and performance indicate that GPT4 performs on par with human annotators. In fact, error analysis and manual examination of annotations suggest the LLMs potentially outperform human annotators in terms of recall (sensitivity), catching annotations which have been missed. However, while recall might be improved in LLMs versus human annotators, precision may suffer in unexpected ways, for example through errors in the use of negation and factuality, even in the case of GPT4. LLMs display a particular tendency to overpredict labels and produce negative annotations in infelicitous contexts, i.e., when humans would deem them irrelevant, creating an amount of noise. However, these negative annotations are not technically incorrect. While accuracy errors could be found in the LLM output, the experts' outputs were not entirely free of them, and previous work by \cite{Van_Veen_2024} suggests LLMs may both be more complete AND more accurate than medical experts. There may still be a difference in the type of accuracy errors produced by LLMs, which will have to be investigated in future research. 

In terms of accuracy at the subcategory level, we were surprised to find GPT4 outperformed human agreement by a large margin in real data (0.47 vs 0.35). We hypothesise this is due to the fact that human annotators display higher subjectivity in their style of annotation at the subcategory level (given the lack of predetermined subcategories) and diverge more between them. LLMs are likely to be more `standard' and generic and thus potentially more in agreement with any given human annotator. More specifically, LLMs tend to be consistent from one annotation to the other with higher recall whereas human annotators showed less consistency. Therefore, if a sentence mentions physical, sexual and emotional abuse, annotators might only mention two out of three but when mentioning all three an LLM is more likely to be in agreement than another annotator, i.e., the LLM will catch more of the perfectly recalled annotations than the second annotator.  

The better performance demonstrated on synthetic data doesn't seem due to LLMs performing better on data they are generating, but rather to the synthetic data being less complex and diverse and thus easier to annotate for both LLMs and humans, as evidenced by GPT4 reaching similar inter-annotator agreement scores to humans (with agreement both in humans and LLM/human 10\% higher for synthetic data). This better performance could still warrant using synthetic data for e.g., training machine learning models (given more reliable labels) but only in cases where the potential loss in diversity is compensated by the increase in label reliability. This will likely depend on the specific application.

\section{Conclusion}
We presented the results of a study examining human and Large Language Models (GPT3.5 and GPT4) performance in extracting mental health factors from adolescent social media data. We performed analyses both on real and synthetic data and found GPT4 performance to be on par with human inter-annotator agreement for both datasets, with substantially better performance on the synthetic dataset. However, we find GPT4 still performing non-human errors in negation and factuality, and synthetic data to be much less diverse and differently distributed than real data. The potential for future applications in healthcare will have to be determined by weighing these factors against the substantial reductions in time and cost achieved through the use of LLMs. 

\section*{Acknowledgment}
I.L., D.W.J., and A.K. are partially supported by the National Institute for Health and Care Research (NIHR) AI Award grant (AI\_AWARD02183) which explicitly examines the use of AI technology in mental health care provision. A.K. declare a research grant from GlaxoSmithKline (unrelated to this work). This research project is supported by the NIHR Oxford Health Biomedical Research Centre (grant NIHR203316). The views expressed are those of the authors and not necessarily those of the UK National Health Service, the NIHR or the UK Department of Health and Social Care.

\bibliography{custom}
\appendix
\section{Appendix A}
\label{appendixA}

Write five Reddit posts from adolescents in subreddits related to mental health and annotate each sentence with the following labels:

- TRAUMA (sexual abuse, physical abuse, emotional abuse, school bullying, death, accident)

- PRECARITY (socioeconomic, parental conflict, parental illness)

- SYMPTOM (self-harm, low self-esteem, anhedonia, panic attack, flashback, psychosis, insomnia)

- CONDITION (eating disorder, depression, bipolar, anxiety, ptsd, adhd, substance abuse/addiction)

- SUICIDALITY 

- TREATMENT

Do not specify the subreddit. Annotate both presence (POSITIVE) and absence (NEGATIVE) of a factor. Here are some examples:

- “I am not feeling suicidal but I can’t sleep at all [SYMPTOM:POSITIVE(insomnia), SUICIDALITY:NEGATIVE]."

- “My sister used to constantly bully me [TRAUMA:POSITIVE(emotional abuse)]."

- “I was harassed for years in secondary school [TRAUMA:POSITIVE(school bullying)]
- “I went on holiday last month [NONE]."

- "My family is quite wealthy [PRECARITY:NEGATIVE(socioeconomic)]."

- "I often cut my wrists with scissors [SYMPTOM:POSITIVE(self-harm)]"

Make sure you always add POSITIVE or NEGATIVE to the factor and specify the subcategory for all factors apart from TREATMENT and SUICIDALITY.

For TREATEMENT and SUICIDALITY you only specify presence or absence, e.g. [TREATMENT:POSITIVE] or [SUICIDALITY:NEGATIVE].

Each post should be 5 sentences in length.

\newpage
\section{Appendix B}
\label{appendixB}

\begin{figure}[h!]
\centering
\includegraphics[width=\textwidth]{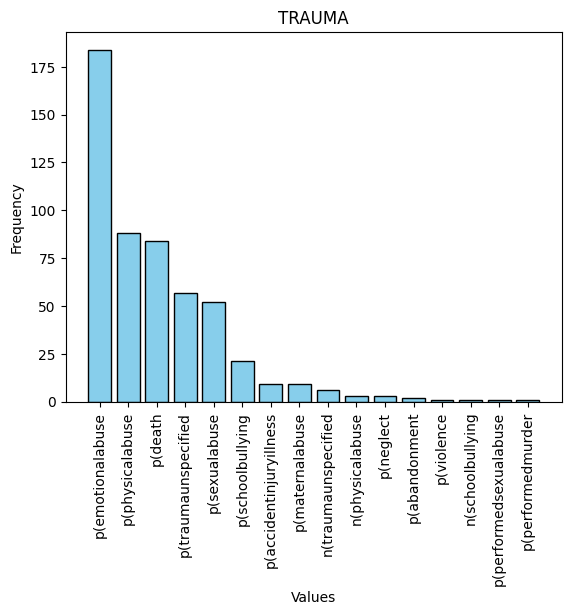}
\caption{Distribution of trauma subcategories (real data)}
\label{fig:trauma}
\end{figure}

\begin{figure}[h!]
\centering
\includegraphics[width=\textwidth]{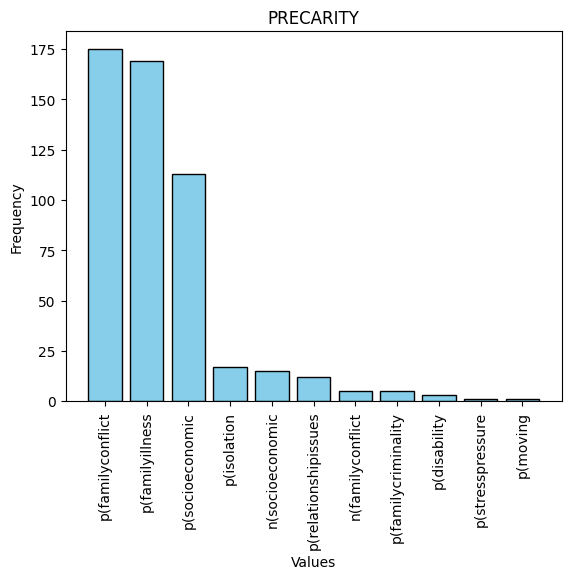}
\caption{Distribution of precarity subcategories (real data)}
\label{fig:precarity}
\end{figure}

\begin{figure}[h!]
\centering
\includegraphics[width=\textwidth]{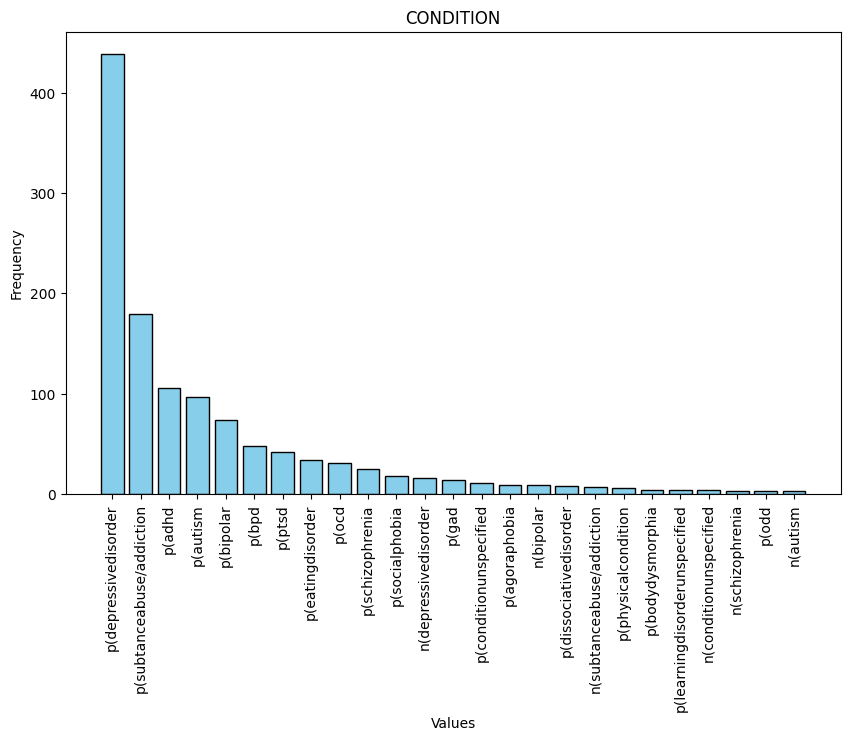}
\caption{Distribution of condition subcategories (real data) (subcategories with n \textgreater 2)}
\label{fig:condition}
\end{figure}

  \begin{figure}[h!]
  \centering
\includegraphics[width=\textwidth]{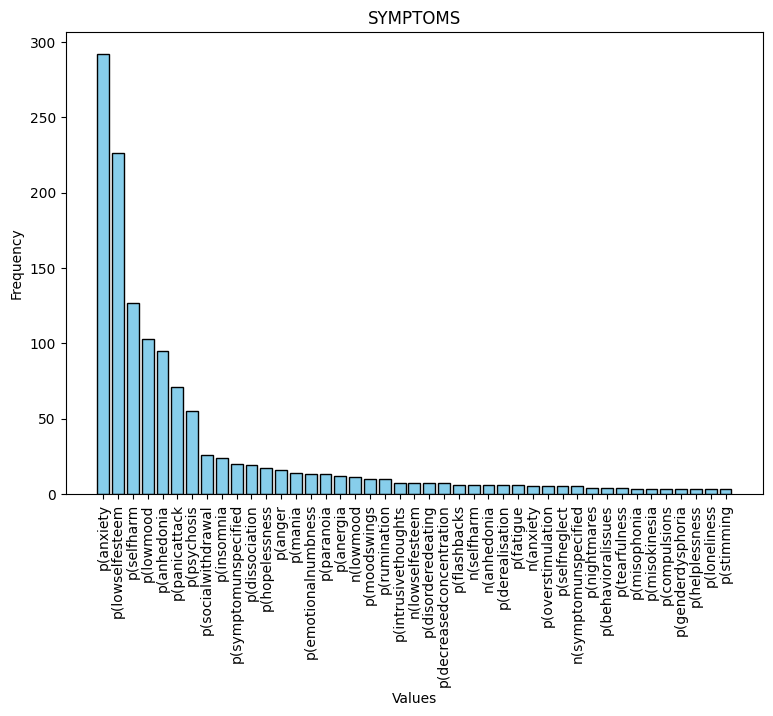}
\caption{Distribution of symptom subcategories (real data) (subcategories with n \textgreater 2)}
\label{fig:symptoms}
\end{figure}

\begin{figure}
    \begin{subfigure}[b]{0.45\textwidth}
        \includegraphics[width=\textwidth]{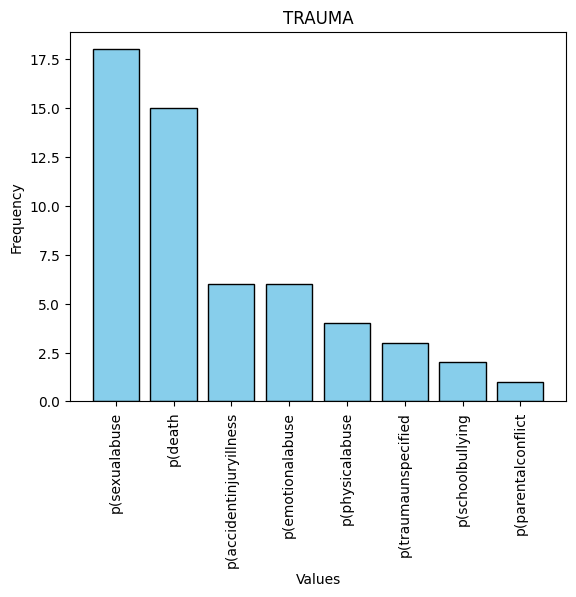}
        \caption{GPT3.5}
        \label{fig:figure1}
    \end{subfigure}
    \hfill
    \begin{subfigure}[b]{0.45\textwidth}
        \includegraphics[width=\textwidth]{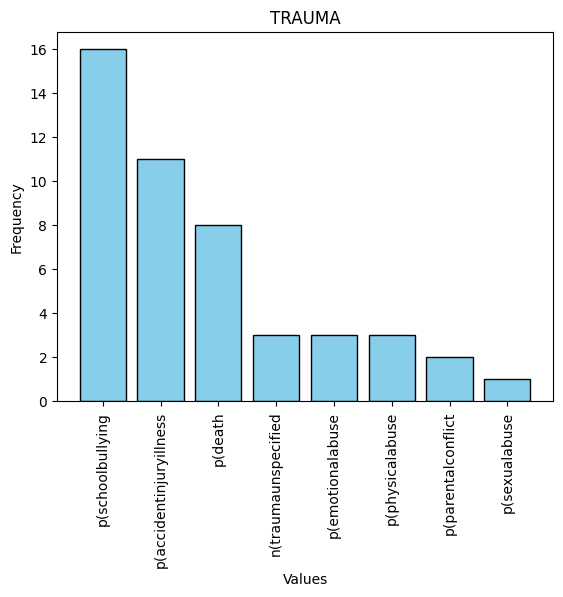}
        \caption{GPT4}
        \label{fig:figure2}
    \end{subfigure}
    \caption{Distribution of trauma subcategories (synthetic data)}
    \label{fig:both}
\end{figure}

\begin{figure}
    \begin{subfigure}[b]{0.45\textwidth}
        \includegraphics[width=\textwidth]{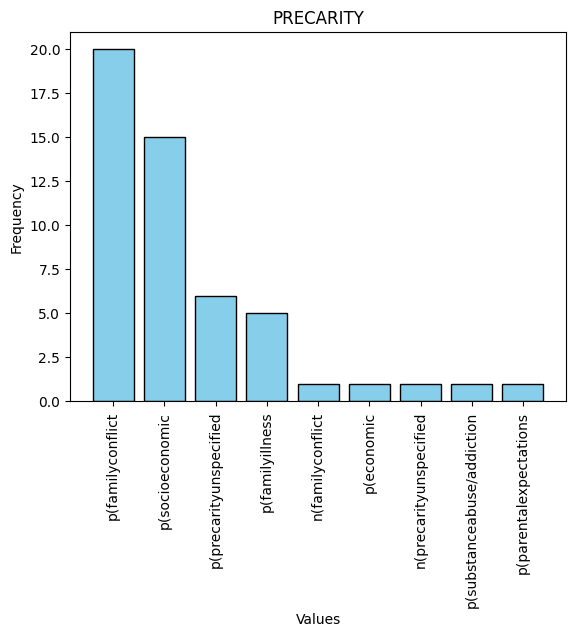}
        \caption{GPT3.5}
        \label{fig:figure1}
    \end{subfigure}
    \hfill
    \begin{subfigure}[b]{0.45\textwidth}
        \includegraphics[width=\textwidth]{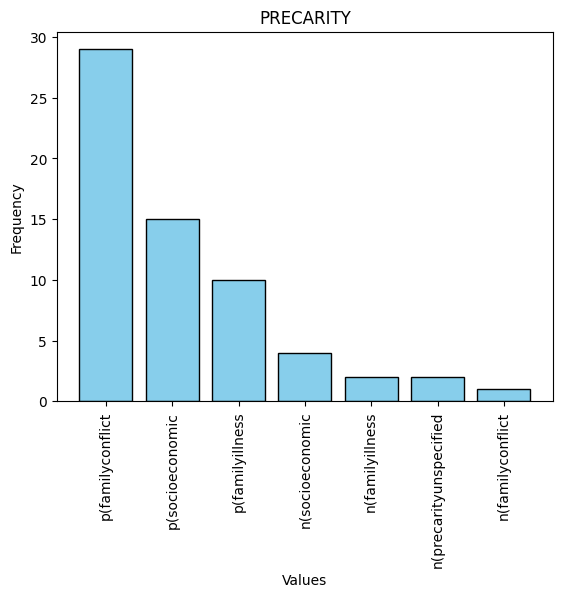}
        \caption{GPT4}
        \label{fig:figure2}
    \end{subfigure}
    \caption{Distribution of precarity subcategories (synthetic data)}
    \label{fig:both}
\end{figure}

\begin{figure}
    \begin{subfigure}[b]{0.45\textwidth}
        \includegraphics[width=\textwidth]{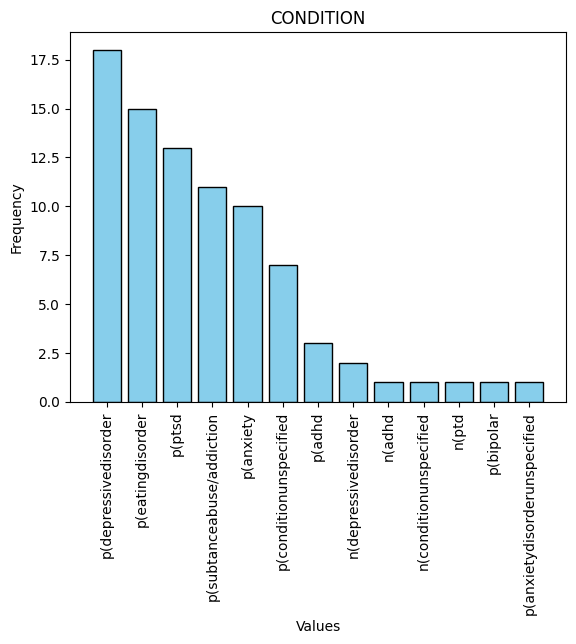}
        \caption{GPT3.5}
        \label{fig:figure1}
    \end{subfigure}
    \hfill
    \begin{subfigure}[b]{0.45\textwidth}
        \includegraphics[width=\textwidth]{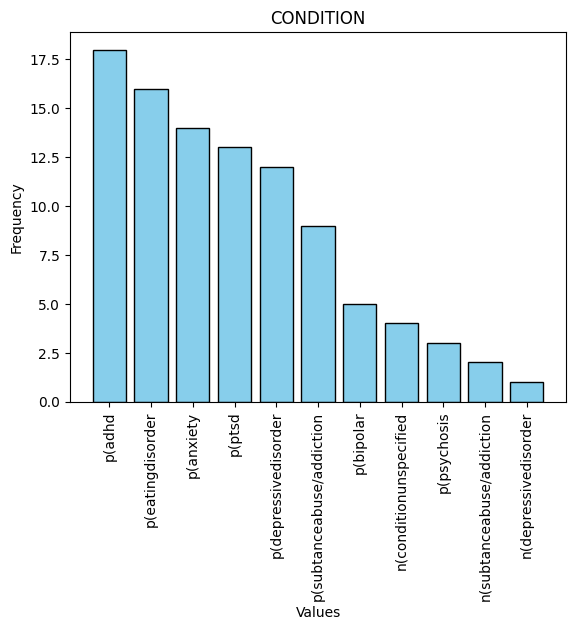}
        \caption{GPT4}
        \label{fig:figure2}
    \end{subfigure}
    \caption{Distribution of condition subcategories (synthetic data)}
    \label{fig:both}
\end{figure}

\begin{figure}
    \begin{subfigure}[b]{0.45\textwidth}
        \includegraphics[width=\textwidth]{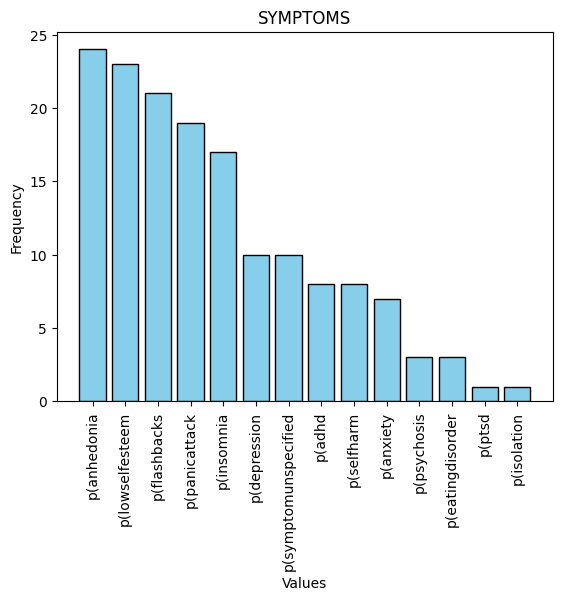}
        \caption{GPT3.5}
        \label{fig:figure1}
    \end{subfigure}
    \hfill
    \begin{subfigure}[b]{0.45\textwidth}
        \includegraphics[width=\textwidth]{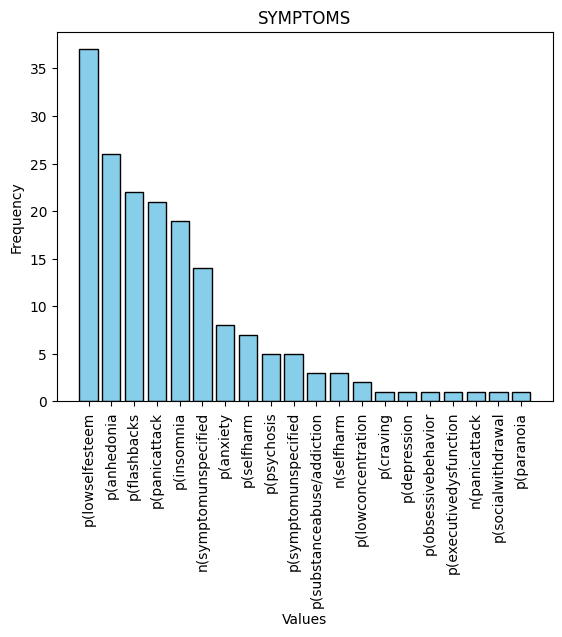}
        \caption{GPT4}
        \label{fig:figure2}
    \end{subfigure}
    \caption{Distribution of symptom subcategories (synthetic data)}
    \label{fig:both}
\end{figure}

\end{document}